\documentclass[runningheads]{llncs}
\usepackage{graphicx}

\usepackage{tikz}
\usepackage{comment}
\usepackage{amsmath,amssymb} 
\usepackage{color}

\usepackage{multirow}
\usepackage{algorithm}
\usepackage{algorithmic}
\usepackage{bbding}
\usepackage{subfigure}
\usepackage{booktabs}

\usepackage[accsupp]{axessibility}  

\begin{document}
\pagestyle{headings}
\mainmatter
\def\ECCVSubNumber{3523}  

\title{Multi-Faceted Distillation of Base-Novel Commonality for Few-shot Object Detection} 

\titlerunning{Multi-Faceted Distillation of Base-Novel
Commonality}
%
\author{Shuang Wu\inst{2} \and Wenjie Pei\inst{2,*} \and  Dianwen Mei\inst{2} \and Fanglin Chen\inst{2} \and Jiandong Tian\inst{3} \and Guangming Lu\inst{1,2,*}} 
\authorrunning{S. Wu et al.}
%
\institute{Guangdong Provincial Key Laboratory of Novel Security Intelligence Technologies \\ \and Harbin Institute of Technology, Shenzhen, China \\
\and Shenyang Institute of Automation, Chinese Academy of Sciences \\
\email{\{wushuang9811, wenjiecoder\}@outlook.com},
\email{\{178mdw, linwers\}@gmail.com},
\email{luguangm@hit.edu.cn},
\email{tianjd@sia.cn}
}

\maketitle

\renewcommand{\thefootnote}{}
\footnotetext{$^*$ Corresponding authors.}

\begin{abstract}
Most of existing methods for few-shot object detection follow the fine-tuning paradigm, which potentially assumes that the class-agnostic generalizable knowledge can be learned and transferred implicitly from base classes with abundant samples to novel classes with limited samples via such a two-stage training strategy. However, it is not necessarily true since the object detector can hardly distinguish between class-agnostic knowledge and class-specific knowledge automatically without explicit modeling. In this work we propose to learn three types of class-agnostic commonalities between base and novel classes explicitly: recognition-related semantic commonalities, localization-related semantic commonalities and distribution commonalities. We design a unified distillation framework based on a memory bank, which is able to perform distillation of all three types of commonalities jointly and efficiently. Extensive experiments demonstrate that our method can be readily integrated into most of existing fine-tuning based methods and consistently improve the performance by a large margin. Code is available at:  \textcolor{blue}{\url{https://github.com/WuShuang1998/MFDC}}.

\keywords{Few-shot; Object Detection; Knowledge Distillation; Commonality}
\end{abstract}

\section{Introduction}
\label{sec:intro}
\begin{figure}[t]
\centering
\includegraphics[width=0.8\linewidth]{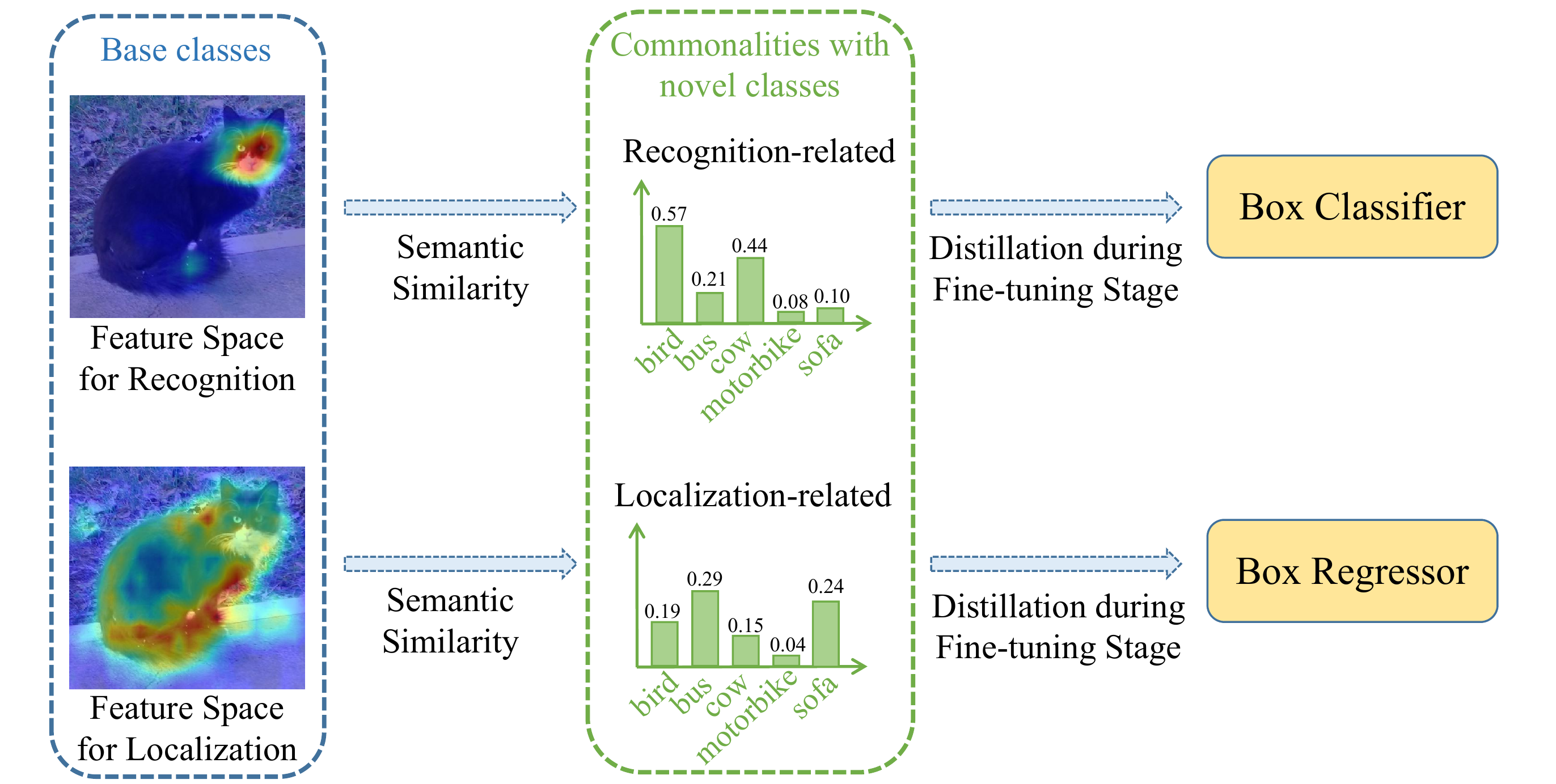}
\caption{Given a cat sample from the base class `Cat', we measure the semantic similarities between it and each of novel classes in the optimized feature space for both object recognition and localization, which are interpreted as the recognition- and localization-related semantic commonalities, respectively. These learned commonalities are distilled during the fine-tuning stage to improve the performance of the object detector on novel classes. Note that the visualizations by Grad-CAM++~\cite{cam2018} show that the learned features for recognition focus on the local salient regions while the localization pays more attention to the global boundary or shape features.}
\label{fig:insight}
\end{figure}

Few-shot object detection aims to learn effective object detectors for novel classes with limited samples, leveraging the generalizable prior knowledge learned from abundant data of base classes. Compared to general object detection~\cite{fast2015,faster2015}, few-shot object detection is supposed to be able to generalize across different classes rather than just across different samples within a class. It is also more challenging than few-shot classification~\cite{maml2017,prototypical2017,matching2016} in that it demands to learn the transferable knowledge not only on recognition, but also on localization.

A prominent modeling paradigm for few-shot object detection is fine-tuning framework~\cite{retentive2021,defrcn2021,fsce2021,tfa2020,mpsr2020}, which first pre-trains the object detector using the samples from base classes, then fine-tunes the model on novel classes. Based on such two-stage training strategy, many methods are proposed to deal with a specific challenge in few-shot object detection, such as MPSR~\cite{mpsr2020} which tackles the problem of scale variation, FSCE~\cite{fsce2021} for alleviating confusion between novel classes, and Retentive R-CNN~\cite{retentive2021} suppressing the performance degradation on base classes during fine-tuning. A potential hypothesis of such fine-tuning paradigm is that the class-agnostic prior knowledge for object detection could be transferred from base classes to novel classes implicitly. Nevertheless, the object detector can hardly distinguish between class-agnostic knowledge and class-specific knowledge automatically without explicit modeling.  

In this work we propose to learn multi-faceted commonalities between base classes and novel classes explicitly in the fine-tuning framework, which is class-agnostic and can be transferred across different classes. Then we perform distillation on the learned commonalities to circumvent the scarcity of novel classes and thereby improve the performance of the object detector on novel classes. To be specific, we aim to learn three types of base-novel commonalities: 1) the recognition-related semantic commonalities like similar appearance features shared among semantically close classes; 2) the localization-related semantic commonalities such as the similar object shape or boundary features between different classes; 3) the distribution commonalities in feature space shared between similar classes like close mean and variance of features in a presumed Gaussian distribution~\cite{one2012}. Consider the example in Figure~\ref{fig:insight}, we first learn the optimized feature spaces for object recognition and localization respectively. Then we measure the semantic similarities between a given cat sample (from the base class `Cat') and each of novel classes in each feature space. The obtained similarity distribution in the feature space for recognition is interpreted as the recognition-related semantic commonalities, and the same applies to the localization-related semantic commonalities. The learned commonalities are further distilled towards their corresponding tasks respectively during fine-tuning of the object detector on novel classes, namely recognition-related commonalities for object classification and localization-related commonalities for object bounding box regression. Consequently, all samples in base classes that share commonalities with a novel class can be leveraged to train the object detector on this novel class, which is equivalent to augment the training data for novel classes. Note that the learned features for recognition and localization focus on different object areas: the recognition captures the local salient regions (e.g., the head of cat in Figure~\ref{fig:insight}) whilst the localization pays more attention to the global boundaries as shown in Figure~\ref{fig:insight}. Thus we decouple the feature spaces for object recognition and localization and learn the corresponding commonalities in the decoupled feature spaces separately. Inspired by Distribution Calibration~\cite{dc2020}, we learn the distribution commonalities by estimating the feature variance for a novel class via reference to the closed base classes, and distill the obtained commonalities by sampling for data augmentation. To conclude, we make following contributions.
\begin{itemize}
    \item We learn three types of generalizable commonalities between base and novel classes explicitly, which can be transferred from base classes to novel classes.
    \item We design a unified distillation framework based on a memory bank, which is able to distill all three types of learned commonalities jointly and efficiently in an end-to-end manner during the fine-tuning stage.
    \item Our method can be integrated into most of fine-tuning based methods. Extensive experiments show that our method leads to substantial improvements when integrated into various classical methods. As a result, our method advances the state-of-the-art performance by a large margin.
\end{itemize}

\section{Related Work}
\label{sec:relatedwork}
\noindent\textbf{Few-Shot Image Classification.}
Few-shot image classification, which aims to recognize novel categories with limited annotated instances, has received increased attention in the recent past. Optimization-based approaches~\cite{maml2017,convex2019,metasgd2017} modify the classical gradient-based optimization for fast adaption to new tasks. Metric-based approaches~\cite{prototypical2017,compare2018,matching2016,deepemd2020} learn a metric space where instances could be recognized by comparing the distance to the prototype of each category. Hallucination-based approaches~\cite{shrinking2017,imaginary2018,dc2020} learn to generate novel samples to deal with data scarcity. Compared to image classification, few-shot object detection which has to consider localization in addition, is still under-explored.

\noindent\textbf{Few-Shot Object Detection.}
Early works of few-shot object detection focus on the meta-learning paradigm~\cite{attentionrpn2020,qafewdet2021,dcnet2021,fsrw2019,tip2021,cme2021,metadet2019,view2020,metarcnn2019,sqmg2021}, which introduces a meta-learner to leverage meta-level knowledge that can be transferred from base classes to novel classes. Recently, researchers find out that the simple fine-tuning based approaches~\cite{fadi2021,retentive2021,fscn2021,defrcn2021,fsce2021,tfa2020,universal2021,svd2021,mpsr2020,halluc2021,srrfsd2021} could outperform most of meta-learning based approaches. TFA~\cite{tfa2020} proposes a two-stage fine-tuning process that only fine-tunes the prediction layer. FSCE~\cite{fsce2021} rescues misclassifications between novel classes by supervised contrastive learning. UP-FSOD~\cite{universal2021} devises universal prototypes to enhance the generalization of object features. Retentive R-CNN~\cite{retentive2021} regularizes the adaptation during fine-tuning to maintain the performance on base classes. DeFRCN~\cite{defrcn2021} proposes to decouple the features for RPN and R-CNN. All these methods learn to detect novel instances by implicitly exploiting the class-agnostic knowledge learned from base classes. Instead, we address few-shot object detection by distilling the multi-faceted commonalities between base classes and novel classes.

\noindent\textbf{Knowledge Distillation.}
Classical knowledge distillation aims at transferring knowledge from a model (teacher) to the other (student). \cite{distilling2015} introduces the soft prediction of the teacher network as dark knowledge for distillation. \cite{fitnets2014} leverages the intermediate representations learned by teacher to guide student. \cite{attentiontransfer2017} proposes to transfer attention information of teacher. Several works~\cite{self2019,progressive2021,distortion2019,regularizing2020,own2019} use the student itself as a teacher, named self-distillation. Inspired by these works, we design a novel distillation framework to distill commonalities between base classes and novel classes based on a memory bank.

\section{Multi-Faceted Distillation of Base-Novel Commonality}
\label{sec:method}
In this section, we start with the preliminary of few-shot object detection, then we introduce our method which distills the multi-faceted base-novel commonalities to circumvent the scarcity of training samples in few-shot object detection.

\subsection{Preliminary}
We follow the standard few-shot object detection settings introduced in~\cite{fsrw2019,tfa2020} and split classes into two sets: base classes $C_b$ with abundant annotated instances, and novel classes $C_n$ with only K (usually less than 30) instances per category. Our proposed method involves the two-stage training procedure~\cite{tfa2020}. In the first stage, the Faster R-CNN~\cite{faster2015} detector is trained with all the available samples of base classes. In the second stage, the pre-trained detector is fine-tuned on samples of both base and novel classes.

\begin{figure}[t]
\centering
\includegraphics[width=1.0\linewidth]{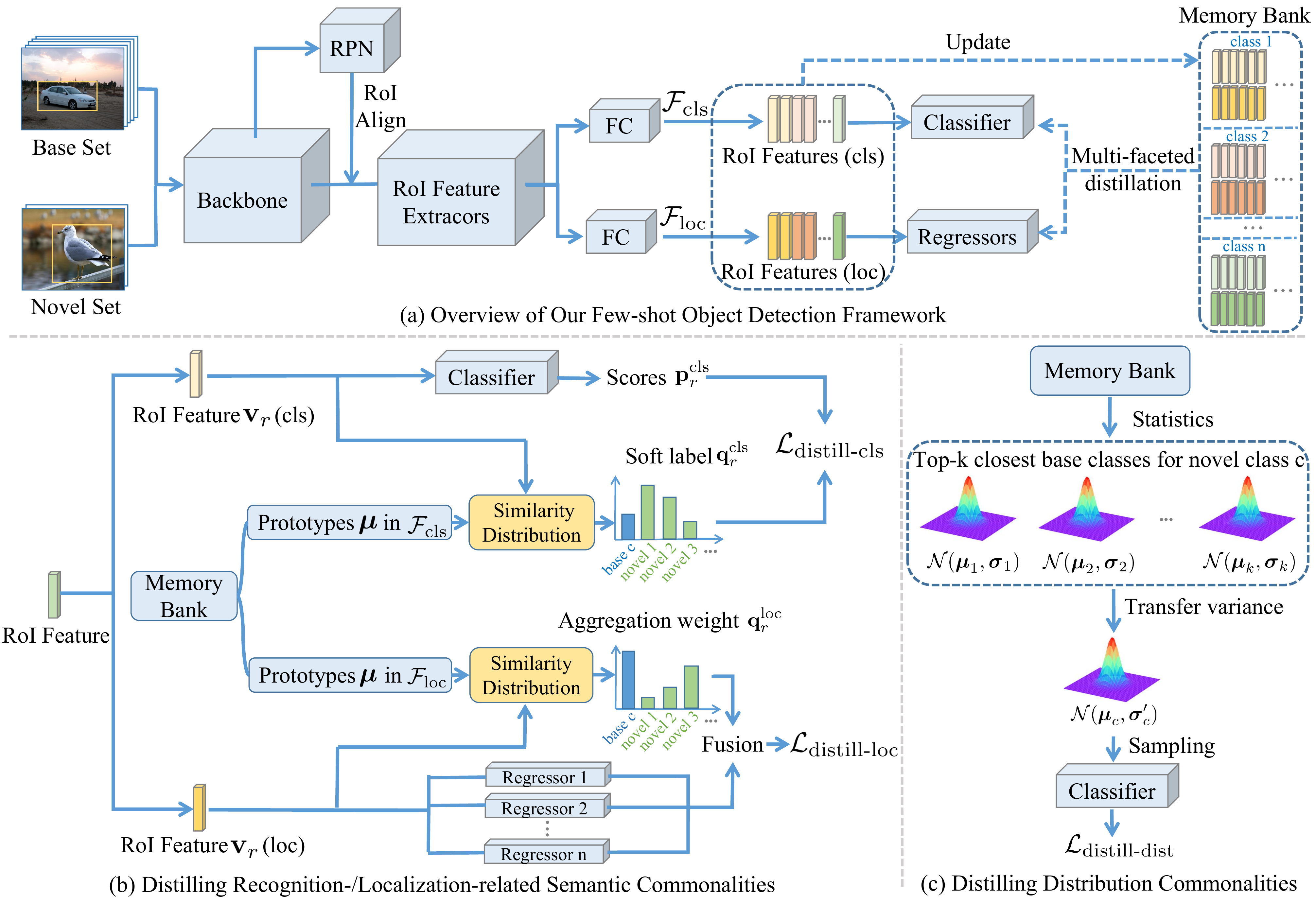}
\caption{The framework of our approach. (a) The RoI features are decoupled into two separate feature spaces for classification  $\mathcal{F}_\text{cls}$ and bounding box regression $\mathcal{F}_\text{loc}$, respectively. During the fine-tuning stage, the recognition-related and distribution commonalities are learned in $\mathcal{F}_\text{cls}$ while the localization-related commonalities are learned in $\mathcal{F}_\text{loc}$. All three types of commonalities are distilled in a unified framework based on a memory bank. (b) The recognition-related commonalities are distilled by viewing them as the soft labels to supervise the classifier whereas the localization-related commonalities are used as aggregation weights to fuse all regressors. (c) We distill the variance for a novel class via reference to the top-$k$ closest base classes, and sample examples from the calibrated distribution to train the classifier.}
\label{fig:framework}
\end{figure}

Different from existing works~\cite{retentive2021,defrcn2021,fsce2021,tfa2020,mpsr2020} that create a small balanced training set with K novel samples and K base samples in the second stage, we fine-tune the detector with abundant samples of base classes which are used in the first stage (the training details are described in the supplementary materials). Thus, we are able to distill the multi-faceted commonalities that can be transferred from abundant samples of base classes to limited samples of novel classes to circumvent the data scarcity. Specifically, we distill three types of base-novel commonalities to learn robust detector for novel classes, including 1) the recognition-related semantic commonalities 2) the localization-related semantic commonalities, and 3) the distribution commonalities. Figure~\ref{fig:framework} illustrates the overall framework of our method.

\subsection{Distilling Recognition-related Semantic Commonalities}
\label{sec:recognition}
Semantically close categories tend to share similar high-level semantic commonalities that is related to object recognition, such as similar appearance between cow and horse. We aim to distill such semantic commonalities between base and novel classes to guide the learning of the object detector on novel classes.

Classical knowledge distillation~\cite{distilling2015} transfers knowledge from a larger teacher model to a student model. The transferred knowledge is represented as the predicted probabilistic distribution on all classes by the teacher model, which can be interpreted as the similarities of current sample to each class. The knowledge distillation is performed by using such probabilistic distribution as the soft labels to supervise the learning of the model together with the one-hot hard labels.

We draw inspiration from such classical way of knowledge distillation but conduct distillation in a different way. To distill the recognition-related semantic commonalities between base and novel classes, we measure the similarities of samples in base classes to each novel class. Since there is no sufficient samples from novel classes for learning a teacher model, we calculate such similarities in a pre-learned feature space $\mathcal{F}_\text{cls}$ directly instead of predicting class probabilities by a teacher model. Formally, given a foreground region proposal $r$ from a base class, which is generated by the region proposal network (RPN), we define the similarity of it to a novel class $c$ as the cosine distance between its RoI feature $\mathbf{v}_r$ and the prototype $\boldsymbol{\mathbf{\mu}}_c$ of the class $c$ in the pre-learned feature space $\mathcal{F}_\text{cls}$:
\begin{equation}
    d_r^{c} = \alpha \cdot \frac{ \mathbf{v}_r^T \boldsymbol{\mu}_c}{\left\| \mathbf{v}_r \right\|  \left\| \boldsymbol{\mu}_c \right\|}, c \in \mathcal{C}_n.
    \label{eqn:simi_novel}
\end{equation}
Herein, $\mathcal{C}_n$ is the set of novel classes and $\alpha > 0$ is the scaling factor. The prototype $\boldsymbol{\mathbf{\mu}}_c$ is obtained by averaging the object features of a candidate set (implemented as a memory bank, will be elaborated on in Section~\ref{sec:framework}) in the novel class $c$:
\begin{equation}
    \boldsymbol{\mu}_c = \frac{1}{n_c}\sum_{i=1}^{n_c}\mathbf{f}_c^i,
  \label{eq:mu}
\end{equation}
where $\mathbf{f}_c^i$ is the vectorial feature for the $i$-th object in the candidate set and $n_c$ is size of the set. Since we focus on distilling the base-novel commonalities to circumvent the scarcity of training samples in novel classes, the base-base commonalities are ignored to allocate all model capacity to base-novel commonalities. As a result, the similarities of a region proposal $r$ from a base class to other base classes are defined as a small constant value:
\begin{equation}
    d_r^c = -\alpha, c \in \mathcal{C}_b \setminus \{c_\text{gt}\},
\end{equation}
where $\mathcal{C}_b$ denotes the set of base classes and $\alpha$ is the same scaling factor as in Equation~\ref{eqn:simi_novel}. Note that we also calculate the cosine similarity between $r$ and its groundtruth class $c_\text{gt}$ following Equation~\ref{eqn:simi_novel} to guarantee the predicting accuracy (w.r.t. $c_\text{gt}$).
Finally we normalize the similarities of sample $r$ to all classes by a softmax function:
\begin{equation}
    \mathbf{q}^{\text{cls}}_{r,c} = \frac{ \text{exp}({ d_r^c})}{\sum_{i=1}^{C}\text{exp}( d_r^i)}, c \in \mathcal{C}_n \cup \mathcal{C}_b.
\label{eqn:soft}
\end{equation}
Assuming that a foreground region proposal $r$ has $0$ commonality with background $c_\text{bg}$, we obtain the complete similarity distribution for $r$: $\mathbf{q}_r^{\text{cls}} = [\mathbf{q}_{r}^{\text{cls}};0]$.

Similar to the classical knowledge distillation, we utilize the obtained similarities of a region proposal as soft labels to supervise the learning of our object detector. In particular, we perform such distillation during the fine-tuning stage of the detector. Formally, for the region proposal $r$ from a base class, we minimize the Kullback-Leibler (KL) divergence between the soft labels $\mathbf{q}_r^{\text{cls}}$ and the predicted class probabilities $\mathbf{p}_r^{\text{cls}}$ by the object detector: 
\begin{equation}
\mathcal{L}_\text{distill-cls} = \sum_{c \in \mathcal{C}_n \cup \mathcal{C}_b \cup \{c_\text{bg}\}} (\mathbf{q}_{r,c}^{\text{cls}} \log \mathbf{q}_{r,c}^{\text{cls}}-\mathbf{q}_{r,c}^{\text{cls}} \log \mathbf{p}_{r,c}^{\text{cls}}).
\label{eq:semantic}
\end{equation}

\noindent\textbf{Rationale.} We learn the semantic commonalities that are related to object recognition by measuring the similarities of samples from base classes to each novel class in a pre-defined feature space. Then the learned commonalities (after normalization) are viewed as soft labels to supervise the fine-tuning of the object detector. Consequently, all samples in base classes that share recognition-related semantics with a novel class can be leveraged to train the object detector on this novel class. In this sense, the proposed commonality distillation significantly augments the training data for novel classes, thereby improving the performance of the object detector on novel classes.

\subsection{Distilling Localization-related Semantic Commonalities}
\label{sec:localization}
Besides the recognition-related semantic commonalities, similar categories also share semantic commonalities that is related to object localization such as similar shape or boundary features. Distilling such commonalities between similar base and novel classes enables the object detector to learn transferable knowledge on localization from abundant base class samples, thereby improving its performance of object detection on novel classes. 

The localization-related semantic commonalities are distilled in a similar way as the recognition-related commonalities in Section~\ref{sec:recognition}. We also learn the localization-related commonalities by measuring the similarities of samples in base classes to each novel class in a pre-learned feature space $\mathcal{F}_\text{loc}$. One of the key differences between distillation of two different types of commonalities (recognition- or localization-related) is that they are learned in different pre-learned feature spaces: each feature space should be learned by optimizing the corresponding task (object classification or localization), as illustrated in Figure~\ref{fig:insight}. We present an efficient implementation in Section~\ref{sec:framework}.

The learned localization-related commonalities is represented as the normalized similarities in the same form shown in Equation~\ref{eqn:soft}. In contrast to viewing the recognition-related commonalities as soft labels for supervision, the localization-related commonalities are leveraged as normalized weights to aggregate all class-specific bounding box regressors for object localization. This is based on the intuition that an object can be localized by not only the bounding box regressor for its groundtruth class, but also the regressors for the similar classes, more similarities leading to more confidence. Formally, given a region proposal $r$ from a base class, its bounding box is predicted as offsets $\mathbf{t}=(t_x,t_y,t_w,t_h)$ to the groundtruth position by aggregating the predictions of all regressors for $C$ classes. Then the detector is optimized by minimizing the error between the aggregated prediction and the groundtruth using the smoothed L1 loss~\cite{fast2015}:
\begin{equation}
    \mathcal{L}_\text{distill-loc}=\sum_{c=1}^C \mathbf{q}_{r,c}^{\text{loc}} \cdot \!\!\!\!\!\! \sum_{i\in\{x,y,w,h\}} \!\!\!\!\!\!\text{Smooth}_{L1}({t}_i^c-u_i),
\label{eq:pre_loc}
\end{equation}
where $\mathbf{q}^{\text{loc}}_{r}$ is the normalized similarities representing the localization-related commonalities. $u_i$ is the bounding-box regression groundtruth for $r$ while ${t}_i^c$ is the prediction of box regressor for the class $c$.

\noindent\textbf{Rationale.} The similarities between samples from base classes to each novel class in a pre-learned feature space $\mathcal{F}_\text{loc}$ towards localization are learned as the localization-related commonalities, and are further used as aggregation weights to fuse regressors for all classes. To be specific, a sample (object) from a base class is localized by referring to the predictions of all regressors for the novel classes sharing localization-related commonalities with this sample. It is equivalent to training these regressors with the sample. As a result, all regressors for novel classes are optimized with a lot of additional training samples from base classes, which yields better performance of localization.

\subsection{Distilling Distribution Commonalities}
\label{sec:distribution}
Semantically similar categories usually follow similar data distributions, such as close mean and variance of features in a presumed Gaussian distribution between these categories~\cite{one2012}. Hence, the third type of commonalities between base and novel classes that we aim to distill is the distribution commonalities. Inspired by Distribution Calibration~\cite{dc2020} in few-shot image classification, we distill the distributional statistics from base classes to calibrate the distribution of those similar novel classes. Consequently, we can sample sufficient examples for these novel classes to improve the performance of the object detector on novel classes.

Unlike Distribution Calibration which transfers both the mean and variance of base classes to novel classes, we only distill the variance of base classes while preserving the mean values of novel classes. This is because transferring both the mean and variance of base classes would result in the distributional overlapping between the base and novel classes, making it harder to distinguish between them during object detection. In contrast, the classification between base and novel classes is not required in the few-shot classification setting. 

Assuming that each feature dimension follows a Gaussian distribution, which is consistent with Distribution Calibration~\cite{dc2020}, we first calculate the mean and variance per feature dimension for both base and novel classes in a pre-learned feature space, and select the top-$k$ semantically closest base classes for each novel class according to the Euclidean distance w.r.t. the mean values (equivalent to the class prototype in Equation~\ref{eq:mu}). Then we can approximate the variance of a novel class using the averaged variance over its top-$k$ closest base classes. Formally, the calibrated variance of a novel class $c$ is estimated by:
\begin{equation}
    \boldsymbol{\sigma}^\prime_c = \frac{1}{k}\sum_{i\in S_c}\boldsymbol{\sigma}_i.
\label{eq:new_sigma}
\end{equation}
Herein, $\boldsymbol{\sigma}_i$ is the variance of the base class $i$ and $S_c$ is the set of top-$k$ closest base classes to the novel class $c$. In this way we are able to sample more examples in this pre-learned feature space for the novel class $c$ following the obtained Gaussian distribution $\mathcal{N}(\boldsymbol{\mu}_c,\boldsymbol{\sigma}^\prime_c)$:
\begin{equation}
    \mathbb{S}_c=\left\{\mathbf{v} | \mathbf{v} \sim \mathcal{N}(\boldsymbol{\mu}_c,\boldsymbol{\sigma}^\prime_c) \right\},
\label{eq:sample}
\end{equation}
where $\boldsymbol{\mu}_c$ is mean of the novel class $c$. $\mathbb{S}_c$ is the set of sampled features, which are further used to train the classifier $f_{\theta}$ of the object detector using the Cross-Entropy loss:
\begin{equation}
\mathcal{L}_\text{distill-dist} = \frac{1}{|\mathbb{S}_c|}\sum_{\mathbf{v}\in \mathbb{S}_c} \text{CE}(c, f_\theta(\mathbf{v})).
\label{eq:dist}
\end{equation}

\subsection{Unified Distillation Framework Based on Memory Bank}
\label{sec:framework}
We propose a unified distillation framework, which is able to distill all three commonalities jointly in an end-to-end manner during the fine-tuning stage. 

Both the recognition-related commonalities and the localization-related commonalities are obtained by calculating the similarities between samples of base classes to each of novel classes in their corresponding (but different) pre-learned feature spaces. Typically such pre-learned feature spaces are independent from the feature space for learning the detector, which is achieved by pre-learning the feature spaces based on other data or other networks. Doing so enables the knowledge distillation between two different feature spaces. However, such implementation has two limitations: 1) the commonalities calculated in the pre-learned feature space may not be accurate since the extracted features for samples of both base and novel classes are potentially not optimized; 2) the whole training is performed in two separated stages, which is not efficient.

We propose to learn the commonalities in the same feature space as that for learning the detector. As shown in Figure~\ref{fig:framework}, we only learn one feature space by the typical feature learning backbone together with the RoI feature extractor based on the training data for current task. Then we decouple the feature space into two separate feature spaces by two projection heads: one (denoted as $\mathcal{F}_\text{cls}$) is connected to the classification head and is used for learning the recognition-related commonalities, the other one (denoted as $\mathcal{F}_\text{loc}$) is connected to the regression head and is used for learning the localization-related commonalities. Each projection head consists of a fully connected layer and a ReLU layer. We first pre-train the detector based on the samples from base classes. Then in the fine-tuning stage, we learn each type of commonalities and perform the commonality distillation jointly in the corresponding feature space. Note that the distribution commonalities are also learned in the feature space $\mathcal{F}_\text{cls}$ since the distribution similarities are intuitively more related to the recognition-related semantics.

\noindent\textbf{Commonality Distillation.} During the fine-tuning of the detector, the feature space is evolving all the time. Thus all types of commonalities are also evolving with the update of the feature space. Meanwhile, the commonality distillation is performed in two aspects. First, the commonalities learned based on the previous training state of feature space are further used to optimize the feature space in the next iteration (state). In this sense, the commonalities are distilled between different training states in the same feature space, which is similar to Self-Knowledge Distillation~\cite{progressive2021}. Second, the recognition-related and distribution commonalities are also distilled from the feature space $\mathcal{F}_\text{cls}$ to the classification head while the localization-related commonalities are distilled from $\mathcal{F}_\text{loc}$ to the localization head, yielding more precise classifier and regressors.

\noindent\textbf{Memory Bank.} During the fine-tuning of the detector, the commonalities are evolving with the update of the feature space. However, calculating the prototype for each class (including base and novel classes) from scratch using all available samples in the training set, which is involved in learning all three types of commonalities, is quite computationally expensive due to the feature extraction for all samples. To address this problem, we maintain a dynamic memory bank to store the features (in both $\mathcal{F}_\text{loc}$ and $\mathcal{F}_\text{cls}$) of a maximum number of $L$ RoI features for each class to improve the efficiency. Denoting the memory bank as $\mathbf{M}=\{\mathbf{m}_c\}^C_{c=1}$ where $C$ is the class number, the RoI features of each class are stored as a queue. During each training iteration, we update the memory bank by enqueuing the current batch of samples to the corresponding class queue and dequeuing the same amount of oldest samples for the same class. Then we can calculate the prototype for each class using the RoI features stored in $\mathbf{M}$. As a result, we do not need to extract features for all samples from scratch each time the feature space is updated, and the operating efficiency is thereby improved significantly. Using memory bank for efficiency has been previously explored in unsupervised learning~\cite{moco2020,bank2018}.

\noindent\textbf{Parameter Learning.}
In the pre-training stage using samples from base classes, we train the object detector with standard Faster R-CNN~\cite{faster2015} losses:
\begin{equation}
\mathcal{L}_\text{det} = \mathcal{L}_\text{rpn} + \mathcal{L}_\text{cls} + \mathcal{L}_\text{reg},
\label{eq:det}
\end{equation}
where $\mathcal{L}_\text{rpn}$ is the loss of the RPN to distinguish foreground from background, $\mathcal{L}_\text{cls}$ is the Cross-Entropy loss for classification, and $\mathcal{L}_\text{reg}$ is the smoothed L1 loss~\cite{fast2015} for the regression of bounding boxes. In the fine-tuning stage, the model is supervised with both the Faster R-CNN loss $\mathcal{L}_\text{det}$ and the losses for the distillation of three types of commonalities, in an end-to-end manner:
\begin{equation}
\mathcal{L} = \mathcal{L}_\text{det} + \lambda_{c}\mathcal{L}_\text{distill-cls} + \lambda_{l}\mathcal{L}_\text{distill-loc} + \lambda_{d}\mathcal{L}_\text{distill-dist},
\label{eq:total_loss}
\end{equation}
where $\lambda_{c}$, $\lambda_{l}$ and $\lambda_{d}$ are hyper-parameters to balance among losses.

\section{Experiments}
\label{sec:exp}
\subsection{Experimental Setup}
\noindent\textbf{Datasets.}
Our approach is evaluated on PASCAL VOC~\cite{voc2010} and MS COCO~\cite{coco2014} datasets. We follow the consistent data construction and evaluation protocol in~\cite{fsrw2019,tfa2020}. For PASCAL VOC, the overall 20 classes are split into 15 base classes and 5 novel classes. We utilize the same three partitions of base classes and novel classes introduced in~\cite{fsrw2019}. All base class instances from PASCAL VOC (07+12) trainval sets are available. Each novel class has $K$ instances available where $K$ is set to 1, 2, 3, 5 and 10. We report AP50 of novel classes (nAP50) on PASCAL VOC 07 test set. For the 80 classes in MS COCO, the 20 classes overlapped with PASCAL VOC are selected as novel classes, the remaining 60 classes are selected as base classes. Similarly, we report COCO-style AP and AP75 of novel classes on COCO 2014 validation set with $K=1,2,3,5,10,30$.

\noindent\textbf{Implementation Details.}
As a plug-and-play module, our approach can be easily integrated into other fine-tuning based methods. we evaluate our approach on four baselines: TFA~\cite{tfa2020}, Retentive R-CNN~\cite{retentive2021}, FSCE~\cite{fsce2021} and DeFRCN~\cite{defrcn2021}. We train the detector with a mini-batch of 16 on 8 GPUs, 2 images per GPU. ResNet-101~\cite{resnet2016} pre-trained on ImageNet~\cite{imagenet2015} is used as the backbone. The maximum queue size $L$ in our memory bank is tuned to be 2048. 
The scaling factor $\alpha$ is tune to be 5. For distribution distillation, we transfer the variance of top $k=2$ base classes, and sample $|\mathbb{S}_c|=10$ instances from the calibrated distribution for novel class $c$ during each iteration. The weights of each loss are tuned to be $\lambda_{c}=0.1$, $\lambda_{l}=1.0$, $\lambda_{d}=0.1$. Moreover, We begin the distillation after 200 iterations in the fine-tuning stage to perform a basic optimization of the feature space on novel classes.

\setlength\tabcolsep{1.0pt}
\begin{table}[t]
  \centering
  \caption{Comparison of different few-shot object detection methods in terms of nAP50 on three PASCAL VOC Novel Split sets.}
    \scriptsize
  \begin{tabular}{l| c c c c c | c c c c c | c c c c c}
    \toprule
    \multirow{2}{*}{Method / Shots}  & \multicolumn{5}{c|}{Novel Split 1} & \multicolumn{5}{c|}{Novel Split 2} & \multicolumn{5}{c}{Novel Split 3} \\
    & 1 & 2 & 3 & 5 & 10 & 1 & 2 & 3 & 5 & 10& 1 & 2 & 3 & 5 & 10 \\
    \midrule
    LSTD~\cite{lstd2018}      &8.2&1.0&12.4&29.1&38.5&11.4&3.8&5.0&15.7&31.0&12.6&8.5&15.0&27.3&36.3\\
    FSRW~\cite{fsrw2019}   &14.8&15.5&26.7&33.9&47.2&15.7&15.3&22.7&30.1&40.5&21.3&25.6&28.4&42.8&45.9   \\
    MetaDet~\cite{metadet2019} &18.9&20.6&30.2&36.8&49.6&21.8&23.1&27.8&31.7&43.0&20.6&23.9&29.4&43.9&44.1   \\
    Meta R-CNN~\cite{metarcnn2019}  &19.9&25.5&35.0&45.7&51.5&10.4&19.4&29.6&34.8&45.4&14.3&18.2&27.5&41.2&48.1 \\
    RepMet~\cite{repmet2019}      &26.1&32.9&34.4&38.6&41.3&17.2&22.1&23.4&28.3&35.8&27.5&31.1&31.5&34.4&37.2\\
    NP-RepMet~\cite{restoring2020}      &37.8&40.3&41.7&47.3&49.4&41.6&43.0&43.4&47.4&49.1&33.3&38.0&39.8&41.5&44.8\\
    TFA w/cos~\cite{tfa2020}         &39.8&36.1&44.7&55.7&56.0&23.5&26.9&34.1&35.1&39.1&30.8&34.8&42.8&49.5&49.8 \\
    MPSR~\cite{mpsr2020}      &41.7&$-$&51.4&55.2&61.8&24.4&$-$&39.2&39.9&47.8&35.6&$-$&42.3&48.0&49.7\\
    HallucFsDet~\cite{halluc2021}  &47.0&44.9&46.5&54.7&54.7&26.3&31.8&37.4&37.4&41.2&40.4&42.1&43.3&51.4&49.6\\
    Retentive R-CNN\cite{retentive2021} &42.4&45.8&45.9&53.7&56.1&21.7&27.8&35.2&37.0&40.3&30.2&37.6&43.0&49.7&50.1\\
    FSCE~\cite{fsce2021}  &44.2&43.8&51.4&61.9&63.4&27.3&29.5&43.5&44.2&50.2&37.2&41.9&47.5&54.6&58.5\\
    FSCN~\cite{fscn2021} &40.7&45.1&46.5&57.4&62.4&27.3&31.4&40.8&42.7&46.3&31.2&36.4&43.7&50.1&55.6\\
    SRR-FSD~\cite{srrfsd2021}  &47.8&50.5&51.3&55.2&56.8&32.5&35.3&39.1&40.8&43.8&40.1&41.5&44.3&46.9&46.4\\
    SQMG~\cite{sqmg2021}&48.6&51.1&52.0&53.7&54.3&41.6&45.4&45.8&46.3&48.0&46.1&51.7&52.6&54.1&55.0\\
    CME~\cite{cme2021} &41.5&47.5&50.4&58.2&60.9&27.2&30.2&41.4&42.5&46.8&34.3&39.6&45.1&48.3&51.5\\
    Dictionary~\cite{svd2021}&46.1&43.5&48.9&60.0&61.7&25.6&29.9&44.8&47.5&48.2&39.5&45.4&48.9&53.9&56.9\\
    FADI~\cite{fadi2021}&50.3&54.8&54.2&59.3&63.2&30.6&35.0&40.3&42.8&48.0&45.7&49.7&49.1&55.0&59.6\\
    UP-FSOD~\cite{universal2021} &43.8&47.8&50.3&55.4&61.7&31.2&30.5&41.2&42.2&48.3&35.5&39.7&43.9&50.6&53.3\\
    QA-FewDet~\cite{qafewdet2021}&42.4&51.9&55.7&62.6&63.4&25.9&37.8&46.6&48.9&51.1&35.2&42.9&47.8&54.8&53.5\\
    DeFRCN~\cite{defrcn2021}&57.0&58.6&64.3&67.8&67.0&35.8&42.7&51.0&54.5&52.9&52.5&56.6&55.8&60.7&62.5\\
    \midrule
    Ours&\textbf{63.4}&\textbf{66.3}&\textbf{67.7}&\textbf{69.4}&\textbf{68.1}&\textbf{42.1}&\textbf{46.5}&\textbf{53.4}&\textbf{55.3}&\textbf{53.8}&\textbf{56.1}&\textbf{58.3}&\textbf{59.0}&\textbf{62.2}&\textbf{63.7}\\
    \bottomrule
  \end{tabular}
  \label{tab:voc}
\end{table}

\setlength\tabcolsep{1.9pt}
\begin{table}[t]
  \centering
  \caption{Few-shot object detection performance on MS COCO.}
    \scriptsize
  \begin{tabular}{l| c c | c c | c c | c c | c c | c c}
    \toprule
    \multirow{2}{*}{Method}  & \multicolumn{2}{c|}{1-shot} & \multicolumn{2}{c|}{2-shot} & \multicolumn{2}{c|}{3-shot} & \multicolumn{2}{c|}{5-shot} & \multicolumn{2}{c|}{10-shot} & \multicolumn{2}{c}{30-shot}\\
    & nAP & nAP75 & nAP & nAP75 & nAP & nAP75 & nAP & nAP75 & nAP & nAP75 & nAP & nAP75\\
    \midrule
    FSRW~\cite{fsrw2019}&$-$&$-$&$-$&$-$&$-$&$-$&$-$&$-$&5.6&4.6&9.1&7.6\\
    SRR-FSD~\cite{srrfsd2021}&$-$&$-$&$-$&$-$&$-$&$-$&$-$&$-$&11.3&9.8&14.7&13.5\\
    FSCE~\cite{fsce2021}&$-$&$-$&$-$&$-$&$-$&$-$&$-$&$-$&11.9&10.5&16.4&16.2\\
    UP-FSOD~\cite{universal2021}&$-$&$-$&$-$&$-$&$-$&$-$&$-$&$-$&11.0&10.7&15.6&15.7\\
    SQMG~\cite{sqmg2021}&$-$&$-$&$-$&$-$&$-$&$-$&$-$&$-$&13.9&11.7&15.9&14.3\\
    CME~\cite{cme2021}&$-$&$-$&$-$&$-$&$-$&$-$&$-$&$-$&15.1&16.4&16.9&17.8\\
    TFA w/cos~\cite{tfa2020}&3.4&3.8&4.6&4.8&6.6&6.5&8.3&8.0&10.0&9.3&13.7&13.4\\
    MPSR~\cite{mpsr2020}&2.3&2.3&3.5&3.4&5.2&5.1&6.7&6.4&9.8&9.7&14.1&14.2\\
    QA-FewDet~\cite{qafewdet2021}&4.9&4.4&7.6&6.2&8.4&7.3&9.7&8.6&11.6&9.8&16.5&15.5\\
    FADI~\cite{fadi2021}&5.7&6.0&7.0&7.0&8.6&8.3&10.1&9.7&12.2&11.9&16.1&15.8\\
    DeFRCN~\cite{defrcn2021}&6.5&6.9&11.8&12.4&13.4&13.6&15.3&14.6&18.6&17.6&22.5&22.3\\
    \midrule
    Ours&\textbf{10.8}&\textbf{11.6}&\textbf{13.9}&\textbf{14.8}&\textbf{15.0}&\textbf{15.5}&\textbf{16.4}&\textbf{17.3}&\textbf{19.4}&\textbf{20.2}&\textbf{22.7}&\textbf{23.2}\\
    \bottomrule
  \end{tabular}
  \label{tab:coco}
\end{table}

\subsection{Comparison with State-of-the-art Methods}
We integrate our method based on DeFRCN~\cite{defrcn2021}, a state-of-the-art method for few-shot object detection, to compare with other latest methods.

\noindent\textbf{Results on PASCAL VOC.}
Table~\ref{tab:voc} shows the results on PASCAL VOC. It can be observed that our approach outperforms other methods in all novel splits with different numbers of training shots. In particular, our method achieves much larger performance gain in extremely low-shot settings. For instance, for novel split 1, our approach surpasses the previously best method by 6.4\% and 7.7\% in 1-shot and 2-shot scenarios, respectively. It is reasonable because the distillation of commonalities plays more important role in fewer-shot settings.

\noindent\textbf{Results on MS COCO.}
Similar performance improvements by our method can be observed on the MS COCO benchmark. As shown in Table~\ref{tab:coco}, our approach consistently outperforms other state-of-the-art methods in all settings although MS COCO is quite challenging. Particularly, for 1-shot scenarios, our approach pushes forward the current state-of-the-art performance from 6.5\% to 10.8\% in nAP. Besides, the improvement from 6.9\% to 11.6\% in nAP75 demonstrates the effectiveness of our approach on localization.

\setlength\tabcolsep{4pt}
\begin{table}[t]
  \centering
  \caption{Performance of integrating our method with different classical methods in term of nAP50 on Novel Split 1 of PASCAL VOC.}
    \scriptsize
  \begin{tabular}{l | c |c c c c c }
    \toprule
    Baseline Method & Ours & 1-shot & 2-shot & 3-shot & 5-shot & 10-shot \\
    \midrule
    \multirow{2}{*}{TFA w/cos~\cite{tfa2020}}& &39.8&36.1&44.7&55.7&56.0\\
    &\checkmark&\textbf{45.2}&\textbf{47.3}&\textbf{50.6}&\textbf{58.2}&\textbf{58.4}\\
    \midrule
    \multirow{2}{*}{Retentive R-CNN~\cite{retentive2021}}& &42.4&45.8&45.9&53.7&56.1\\
    &\checkmark&\textbf{47.8}&\textbf{48.1}&\textbf{51.4}&\textbf{58.2}&\textbf{58.9}\\
    \midrule
    \multirow{2}{*}{FSCE~\cite{fsce2021}} &&44.2&43.8&51.4&61.9&63.4\\
    &\checkmark&\textbf{48.0}&\textbf{51.6}&\textbf{55.3}&\textbf{63.8}&\textbf{66.2}\\
    \midrule
    \multirow{2}{*}{DeFRCN~\cite{defrcn2021}} &&57.0&58.6&64.3&67.8&67.0\\
    &\checkmark&\textbf{63.4}&\textbf{66.3}&\textbf{67.7}&\textbf{69.4}&\textbf{68.1}\\
    \bottomrule
  \end{tabular}
  \label{tab:baseline}
\end{table}

\subsection{Integration with Different Baseline Methods}
We further integrate our method with different baseline methods to evaluate the robustness of our method. Table~\ref{tab:baseline} presents the performance of four different baselines and our method on Novel Split 1 of PASCAL VOC. Our method consistently boosts the performance distinctly. For instance, when integrated with TFA w/cos~\cite{tfa2020}, our method achieves substantial performance gains: $5.4\%$, 11.2\%, 5.9\%, 2.5\% and 2.4\% from 1-shot to 10-shot respectively. These results reveal the strong robustness of our approach on different baseline methods.

\subsection{Ablation Studies}
\label{sec:ablation}
In this section, we conduct ablation studies by integrating our method with DeFRCN~\cite{defrcn2021}. All experiments are performed on Novel Split 1 of PASCAL VOC. Note that more ablation studies on other hyper-parameters are provided in the supplementary materials.

\noindent\textbf{Effectiveness of each type of commonality.}
Table~\ref{tab:components} shows the effectiveness of each type of commonality. Compared with the baseline in the first line, each individual type of commonality improves the performance distinctly. Combining all three types of commonalities achieves larger performance gain than any individual one. 

\begin{table}[!t]
\begin{minipage}[!t]{0.47\linewidth}
\setlength\tabcolsep{3pt}
\centering
\caption{Effectiveness of each type of commonality. `Recog', `Local', `Dist' refer to the recognition-related, localization-related and distribution commonalities, respectively.}
    \scriptsize
  \begin{tabular}{c c c| c c c}
    \toprule
    \multirow{2}*{\shortstack{Recog}}& \multirow{2}*{\shortstack{Local}}& \multirow{2}*{\shortstack{Dist}} &  \multicolumn{3}{c}{nAP50} \\
    & & & 1-shot & 2-shot & 3-shot \\
    \midrule
    &&&58.5&62.6&65.4\\
    $\checkmark$&&&62.3&64.8&67.3\\
    &$\checkmark$&&59.9&64.1&65.7 \\
    &&$\checkmark$&62.6&65.1&66.2 \\
    $\checkmark$ & & $\checkmark$ & 63.2 & 65.9 & 67.7 \\
    $\checkmark$&$\checkmark$&&62.8&65.6&67.2 \\
    $\checkmark$ & $\checkmark$ & $\checkmark$ &\textbf{63.4}&
    \textbf{66.3}&\textbf{67.7}\\
    \bottomrule
  \end{tabular}
  \label{tab:components}
\end{minipage}
\qquad
\begin{minipage}[!t]{0.46\linewidth}
\setlength\tabcolsep{2.4pt}
\centering
\caption{Effect of using different feature spaces from the object detector to learn commonalities. `Independent' denotes the feature space pre-optimized on ImageNet, and 'uniform' denotes the same feature space as the object detector.}
    \scriptsize
  \begin{tabular}{l| c c c}
    \toprule
    \multirow{2}{*}{Feature space} &  \multicolumn{3}{c}{nAP50} \\
      & 1-shot & 2-shot & 3-shot  \\
    \midrule
    Baseline &58.5&62.6&65.4\\
    Independent&59.6&63.8&66.1\\
    Uniform (ours)&\textbf{62.3}&\textbf{64.8}&\textbf{67.3}\\
    \bottomrule
  \end{tabular}
  \label{tab:space}
\end{minipage}
\end{table}

\noindent\textbf{Learning commonalities in an independent feature space from the object detector.}
Our method learns commonalities in the same (uniform) feature space as the object detector, which allows our model to 1) achieve more accurate commonalities due to more optimized features for current data and 2) perform commonality distillation in an end-to-end manner.
To validate the first merit, we conduct experiments to learn commonalities in an independent feature space from the object detector, which is pre-optimized on ImageNet dataset. All class prototypes and cosine similarities for learning commonalities are calculated in this independent feature space. The results in Table~\ref{tab:space} show that the performance gain in such way is smaller than that of using the same space feature as the object detector (denoted as `Uniform').

\begin{figure}[!t] \centering
\subfigure[] { \label{fig:vis1}
\includegraphics[width=0.47\columnwidth]{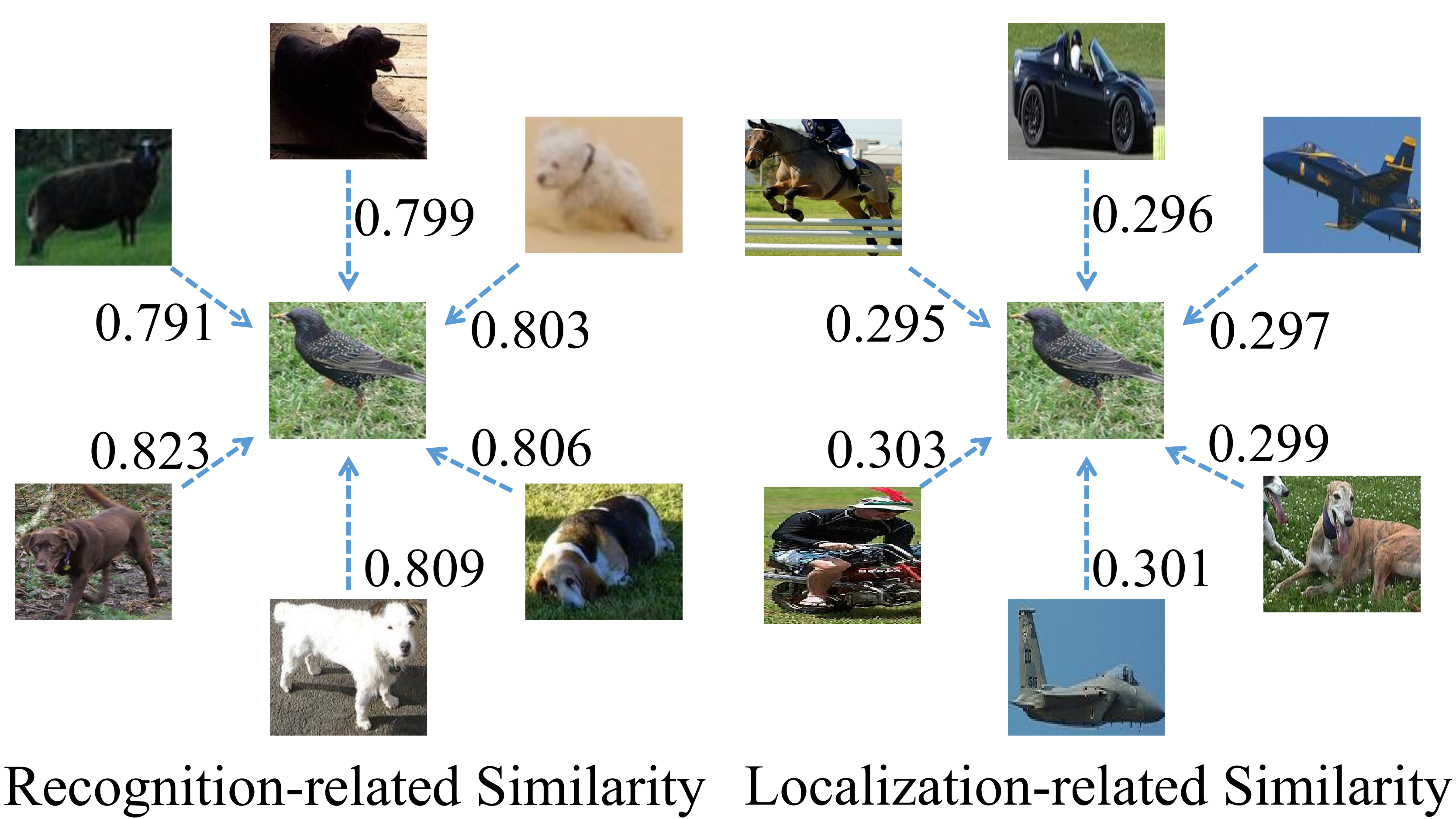}
}
\subfigure[] { \label{fig:vis2}
\includegraphics[width=0.48\columnwidth]{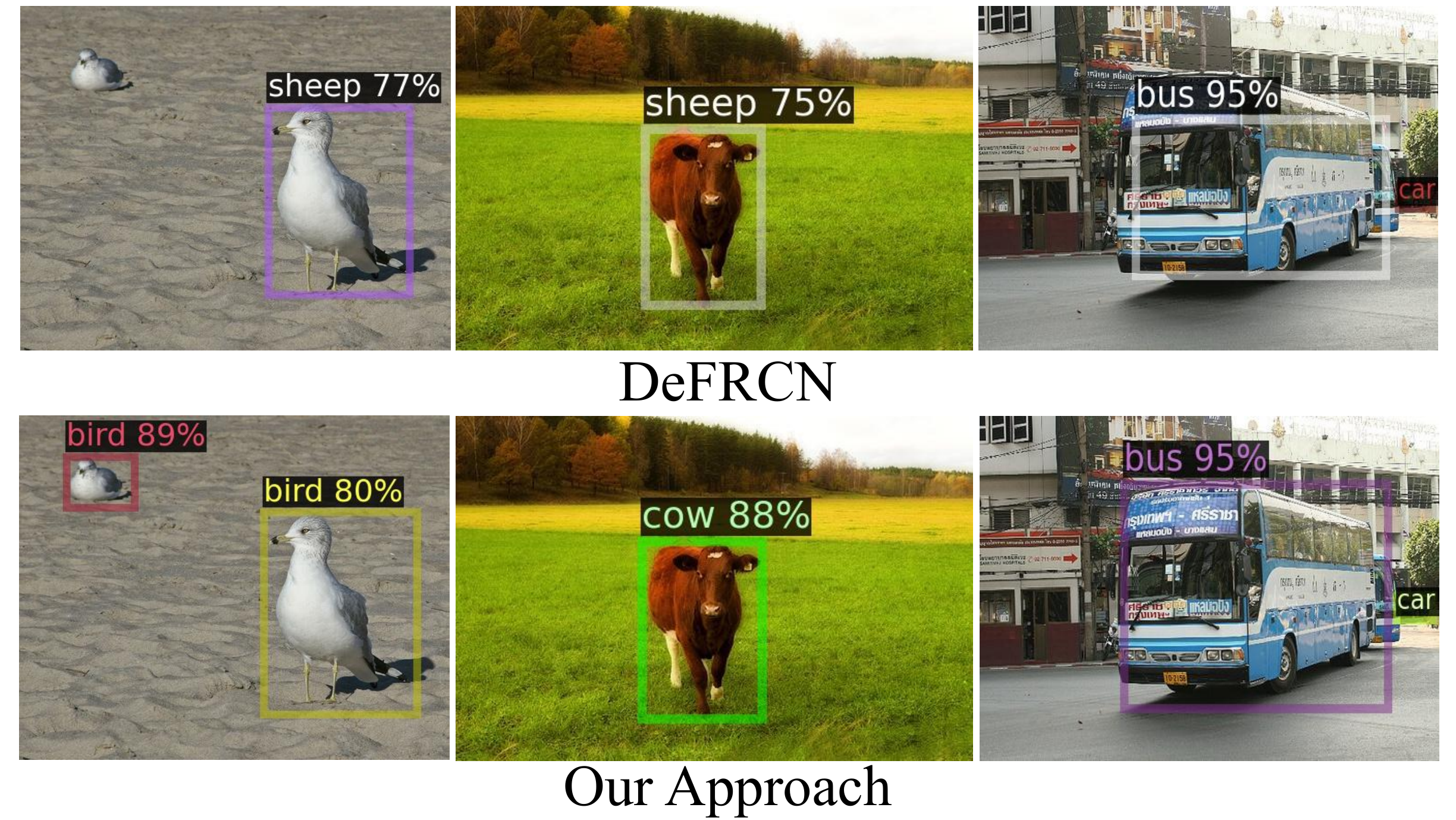}
}
\caption{(a) Visualization of base instances with highest recognition-related similarity and localization-related similarity to the novel class `Bird'. (b) 1-shot object detection results of randomly selected test samples by DeFRCN~\cite{defrcn2021} and our approach on PASCAL VOC Novel Split 1. More examples can be found in the supplementary materials.}
\label{fig:vis}
\end{figure}

\noindent\textbf{Qualitative evaluation.}
To have a qualitative evaluation, we visualize the instances from base classes that have most recognition- and localization-related commonalities (interpreted as semantic similarities) with the novel class `Bird' respectively in Figure~\ref{fig:vis1}. The instances from the semantically similar base classes to `Bird', such as `Dog' and `sheep', tend to have more recognition-related commonalities with `Bird' than other base classes. In contrast, instances from the base classes bearing more shape similarities to `Bird', like `Plane', have more localization-related commonalities with `Bird' than other classes. Such observations are consistent with the different attention distributions in feature space between recognition and localization shown in Figure~\ref{fig:insight}. 
By distilling the multi-faceted commonalities, our object detector is able to perform recognition and localization more accurately, as shown in Figure~\ref{fig:vis2}.

\section{Conclusion}
In this paper, we propose the multi-faceted distillation for few-shot object detection. The key insight is to learn three types of commonalities between base and novel classes explicitly: recognition-related semantic commonalities, localization-related
semantic commonalities and distribution commonalities. Then these commonalities are distilled during the fine-tuning stage based on the memory bank. Our method improves the state-of-the-art performance of few-shot object detection by a large margin.

\subsubsection{Acknowledgements}
This work was supported in part by the NSFC fund (U2013210, 62006060, 62176077), in part by the Guangdong Basic and Applied Basic Research Foundation under Grant (2019Bl515120055, 2021A1515012528, 2022A1515010306), in part by the Shenzhen Key Technical Project under Grant 2020N046, in part by the Shenzhen Fundamental Research Fund under Grant (JCYJ20210324132210025), in part by the Shenzhen Stable Support Plan Fund for Universities (GXWD20201230155427003-20200824125730001, GXWD202012
30155427003-20200824164357001), in part by CAAI-Huawei MindSpore Open Fund(CAAIXSJLJJ-2021-003B), in part by the Medical Biometrics Perception and Analysis Engineering Laboratory, Shenzhen, China, and in part by the Guangdong Provincial Key Laboratory of Novel Security Intelligence Technologies (2022B1212010005).

\clearpage
%
%
\bibliographystyle{splncs04}
\bibliography{egbib}

\clearpage

\title{Supplementary Material for Multi-Faceted Distillation of Base-Novel Commonality for Few-shot Object Detection} 
\titlerunning{Multi-Faceted Distillation of Base-Novel
Commonality}
\author{} 
\authorrunning{S. Wu et al.}
\institute{}

\maketitle

\appendix
\section{More Implementation Details}
Unlike previous works~\cite{defrcn2021,fsce2021,tfa2020} that fine-tune the detector on a small balanced training set with K novel instances and K randomly sampled base instances, we utilize more base instances used in the first stage for fine-tuning, in that our approach requires abundant base instances stored in the memory bank to calculate the prototypes and distributions. Specifically, the training data for fine-tuning consists of two sets: base set with abundant instances and novel set with K instances per class. During each iteration, a batch is composed of two equally sized parts, one from the base set and another from the novel set. Then we update the memory bank by enqueuing the RoI features of instances in current batch to the corresponding class queue. The dimension of RoI features stored in the memory bank is 2048 for DeFRCN~\cite{defrcn2021} baseline and 1024 for other three baselines (TFA~\cite{tfa2020}, Retentive R-CNN~\cite{retentive2021}, FSCE~\cite{fsce2021}). All other training settings (batch size, training iterations, learning rate, etc) are the same as that in corresponding baselines.

\section{Performance for Base Classes}
The proposed commonality distillation from base classes to novel classes allows leveraging the samples of base classes to train the object detector on the novel classes. Such commonality distillation pushes the model to fit the base samples to other classes (novel classes) with semantic similarities instead of their own groundtruth classes (base classes). Thus, it would not lead to the overfitting on the base classes. Table~\ref{tab:base_novel} shows that while the commonality distillation improves the performance of our model on novel classes substantially, it does not degrade the performance on base classes.

\setlength\tabcolsep{4pt}
\begin{table}[ht]
    \scriptsize
    \centering
    \caption{Performance for base classes (bAP50) and novel classes (nAP50) on PASCAL VOC Novel Split 1.}
    \begin{tabular}{l| c c c | c c c}
        \toprule
        \multirow{2}{*}{Method / Shots}  & \multicolumn{3}{c|}{bAP50} & \multicolumn{3}{c}{nAP50} \\
        & 1 & 2 & 3 & 1 & 2 & 3\\
        \midrule
        w/o distillation&78.4&\textbf{78.1}&76.8&57.0&58.6&64.3\\
        w/ distillation&\textbf{78.5}&78.0&\textbf{78.3}&\textbf{63.4}&\textbf{66.3}&\textbf{67.7}\\
        \bottomrule
    \end{tabular}
    \label{tab:base_novel}
\end{table}

\section{Additional Ablation Studies}

\begin{figure}[ht]
\centering
\includegraphics[width=0.96\linewidth]{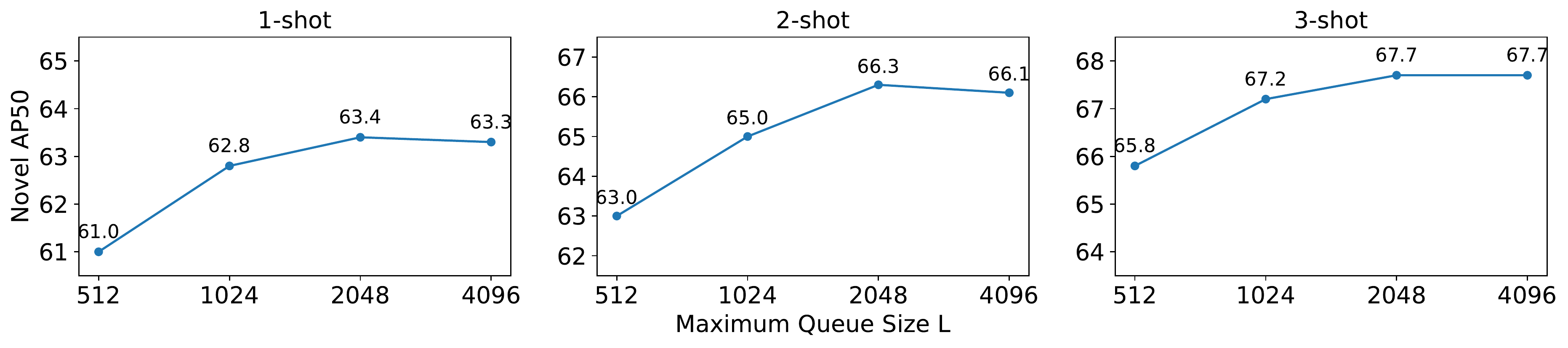}
\caption{Effect of varying the maximum queue size $L$ in the memory bank.}
\label{fig:maxsize}
\end{figure}

\noindent\textbf{Effect of varying the maximum queue size $L$ in the memory bank.} Fig~\ref{fig:maxsize} shows the performance as a function of maximum queue size $L$ in the memory bank for different number of training shots. It can be seen that the performance improves initially as $L$ increases because larger size of queue leads to more accurate estimation of the class prototypes. The performance reaches a plateau at $L=2048$, which is selected for our method in other experiments.

\noindent\textbf{Effect of varying the scaling factor.}
We explore the effect of different scaling factors $\alpha$ for computing similarity distribution and report the results in Table~\ref{tab:scaling}. It can be observed $\alpha=5$ outperforms the other scaling factors for both recognition-related similarity and localization-related similarity. Therefore, we adopt $\alpha=5$ in all of our experiments.
\setlength\tabcolsep{9.2pt}
\begin{table}[ht]
\centering
\scriptsize
\caption{Effect of varying the scaling factor for computing recognition-related and localization-related similarity on PASCAL VOC Novel Split 1.}
\subtable[For recognition-related similarity.]{
  \begin{tabular}{c| c c c}
    \toprule
    \multirow{2}{*}{$\alpha$} &  \multicolumn{3}{c}{nAP50} \\
      & 1-shot & 2-shot & 3-shot  \\
    \midrule
    1&58.4&60.9&62.2\\
    3&60.8&62.5&65.9\\
    5&\textbf{62.3}&\textbf{64.8}&\textbf{67.3}\\
    10&61.3&64.5&66.9\\
    \bottomrule
  \end{tabular}
}
\qquad
\subtable[For localization-related similarity.]{
  \begin{tabular}{c| c c c}
    \toprule
    \multirow{2}{*}{$\alpha$} &  \multicolumn{3}{c}{nAP50} \\
      & 1-shot & 2-shot & 3-shot  \\
    \midrule
    1&59.4&62.3&65.4\\
    3&59.6&63.6&65.4\\
    5&\textbf{59.9}&\textbf{64.1}&\textbf{65.7}\\
    10&58.9&63.6&\textbf{65.7}\\
    \bottomrule
  \end{tabular}
}
\label{tab:scaling}
\end{table}

\noindent\textbf{Effect of varying the loss weights.}
We conduct experiments to evaluate the effect of the hyper-parameters $\lambda_c$, $\lambda_l$ and $\lambda_d$, which control the weight of each distillation loss. As shown in
Table~\ref{tab:lambda}, we obtain the best results with $\lambda_c=0.1$, $\lambda_l=1.0$ and $\lambda_d=0.1$, which are used for all other experiments.
\setlength\tabcolsep{2pt}
\begin{table}[ht]
\centering
\scriptsize
\caption{Effect of varying the weight of each distillation loss on PASCAL VOC Novel Split 1.}
\subtable[Parameter: $\lambda_c$.]{
  \begin{tabular}{c| c c c}
    \toprule
    \multirow{2}{*}{$\lambda_c$} &  \multicolumn{3}{c}{nAP50} \\
      & 1-shot & 2-shot & 3-shot  \\
    \midrule
    0.001&59.2&63.0&64.9\\
    0.01&59.3&62.8&65.9\\
    0.1&\textbf{62.3}&\textbf{64.8}&\textbf{67.3}\\
    1.0&60.7&62.2&63.3\\
    \bottomrule
  \end{tabular}
}
\qquad
\subtable[Parameter: $\lambda_l$.]{
  \begin{tabular}{c| c c c}
    \toprule
    \multirow{2}{*}{$\lambda_l$} &  \multicolumn{3}{c}{nAP50} \\
      & 1-shot & 2-shot & 3-shot  \\
    \midrule
    0.01&58.1&62.0&64.7\\
    0.1&57.6&63.9&65.0\\
    1.0&\textbf{59.9}&\textbf{64.1}&\textbf{65.7}\\
    2.0&59.7&62.6&64.7\\
    \bottomrule
  \end{tabular}
}
\qquad
\subtable[Parameter: $\lambda_d$.]{
  \begin{tabular}{c| c c c}
    \toprule
    \multirow{2}{*}{$\lambda_d$} &  \multicolumn{3}{c}{nAP50} \\
      & 1-shot & 2-shot & 3-shot  \\
    \midrule
    0.001&57.9&63.2&65.9\\
    0.01&60.5&63.4&65.9\\
    0.1&\textbf{62.6}&\textbf{65.1}&\textbf{66.2}\\
    1.0&58.0&58.7&63.2\\
    \bottomrule
  \end{tabular}
}
\label{tab:lambda}
\end{table}

\noindent\textbf{Hyper-parameters for distribution commonalities.}
We study the hyper-parameters, i.e., $k$ and $|\mathbb{S}_c|$ adopted in distribution commonalities. $k$ is the number of the closest base classes to novel class $c$ for transferring the variance. $|\mathbb{S}_c|$ is the number of instances sampled from the calibrated distribution for novel class $c$ during each iteration. As shown in Table~\ref{tab:topk_num}, these two hyper-parameters have a mild impact on the performance, and we observe that $k=2$ and $|\mathbb{S}_c|=10$ work best for nAP50.
\setlength\tabcolsep{6pt}
\begin{table}[ht]
  \centering
  \caption{Ablation study for distribution commonalities. Results (nAP50) are reported on 1-shot of PASCAL VOC Novel Split 1.}
    \scriptsize
  \begin{tabular}{c| c c c c}
    \toprule
    \multirow{2}{*}{\shortstack{Number of the \\ Closest Base Classes}} &  \multicolumn{4}{c}{$|\mathbb{S}_c|$} \\
      & 1 & 5 & 10 & 20\\
    \midrule
    $k=1$&61.3&61.1&61.6&62.4\\
    $k=2$&61.4&61.8&\textbf{62.6}&61.6\\
    $k=3$&62.3&62.3&61.8&60.0\\
    \bottomrule
  \end{tabular}
  \label{tab:topk_num}
\end{table}

\noindent\textbf{`Variance' vs `mean' \& `variance' for distribution commonality.} 
In contrast to Distribution Calibration~\cite{dc2020} transferring both the mean and variance from base classes to novel classes, our method only distills the variance as the distribution commonalities to avoid the distributional overlapping between base and novel classes. We conduct experiments to compare such two mechanisms. The results in Table~\ref{tab:dist_policies} show that transferring both the mean and variance degrades the performance by a large margin than transferring only variance, and performs even worse than the baseline without commonality distillation. 
\setlength\tabcolsep{6pt}
\begin{table}[ht]
    \centering
    \caption{Effect of transfer `mean' for distribution commonality.}
    \scriptsize
    \begin{tabular}{l| c c c}
    \toprule
    \multirow{2}{*}{Dist} &  \multicolumn{3}{c}{nAP50} \\
      & 1-shot & 2-shot & 3-shot  \\
    \midrule
    Baseline&58.5&62.6&65.4\\
    Mean \& variance&57.5&62.4&64.1\\
    Variance&\textbf{62.6}&\textbf{65.1}&\textbf{66.2}\\
    \bottomrule
  \end{tabular}
  \label{tab:dist_policies}
\end{table}

\section{Results over Multiple Runs}
We report the few-shot object detection results (nAP50) over 10 random runs on PASCAL VOC Novel Split 1 in Table~\ref{tab:multi}. It can be observed that our method outperforms the baseline (DeFRCN) under all settings, which shows the effectiveness of our method.
\setlength\tabcolsep{4pt}
\begin{table}[ht]
  \centering
  \caption{Results (nAP50) over 10 random runs on VOC Novel Split 1.}
  \scriptsize
  \begin{tabular}{ l | c c c c c }
    \toprule
    \multirow{2}{*}{Method / Shots} & \multicolumn{5}{c}{nAP50} \\
     & 1 & 2 & 3 & 5 & 10 \\
    \midrule
    DeFRCN&43.8&57.5&61.4&65.3&67.0\\
    Ours&\textbf{53.7}&\textbf{64.3}&\textbf{66.6}&\textbf{69.6}&\textbf{70.4}\\
    \bottomrule
  \end{tabular}
  \label{tab:multi}
\end{table}

\section{More Qualitative Visualizations}
\begin{figure}[ht]
  \centering
   \includegraphics[width=1.0\linewidth]{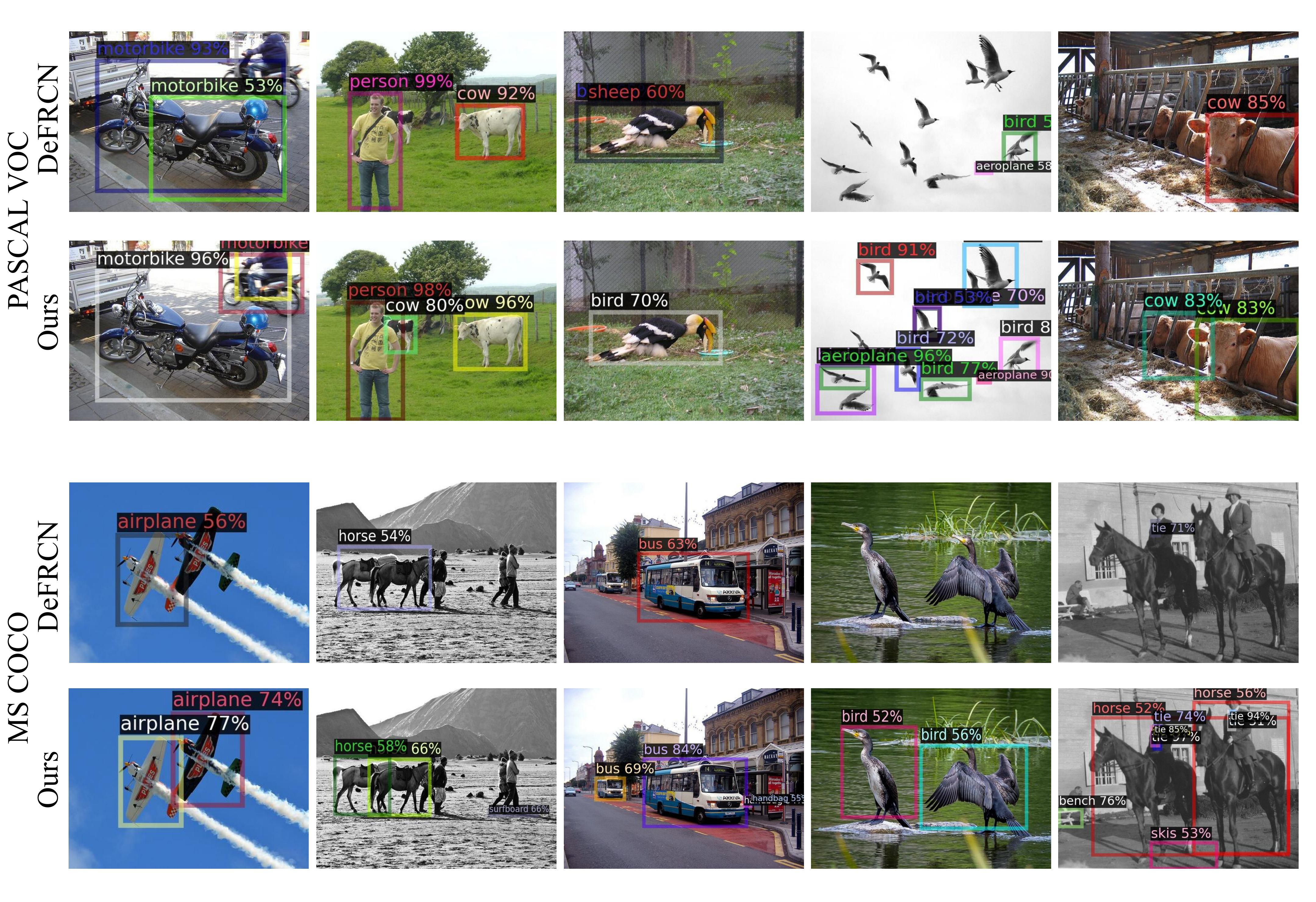}
   \caption{The visualization results of DeFRCN and our approach under 1-shot setting of PASCAL VOC Novel Split 1, and under 1-shot setting of MS COCO.} 
   \label{fig:more_vis}
\end{figure}

In this section, we provide more qualitative visualizations on PASCAL VOC and MS COCO datasets. As shown in Figure~\ref{fig:more_vis}, our approach could rescue various error cases, including missing detections, misclassifications and imprecise localizations.

\end{document}